\documentclass[iicol,pdflatex,sn-mathphys-num]{sn-jnl}

\usepackage{graphicx}%
\usepackage{multirow}%
\usepackage{amsmath,amssymb,amsfonts}%
\usepackage{amsthm}%
\usepackage{mathrsfs}%
\usepackage[title]{appendix}%
\usepackage{xcolor}%
\usepackage{textcomp}%
\usepackage{manyfoot}%
\usepackage{booktabs}%
\usepackage{algorithm}%
\usepackage{algorithmicx}%
\usepackage{algpseudocode}%
\usepackage{listings}%

\usepackage{wrapfig}
\usepackage{multicol}
\usepackage{enumerate}
\usepackage{subcaption}
\usepackage{enumitem}
\usepackage{adjustbox}

\usepackage{chngcntr}
\counterwithin{table}{section}
\counterwithin{figure}{section}
\counterwithin{equation}{section}

\theoremstyle{thmstyleone}%
\theoremstyle{thmstyletwo}%

\theoremstyle{thmstylethree}%

\begin{document}

\title[Article Title]{Class-frequency Guided Noise Schedule for Diffusion Models}


\author*[1]{\fnm{Jiequan} \sur{Cui}}\email{jiequancui@gmail.com}

\author[2]{\fnm{Beier} \sur{Zhu}} 

\author[2]{\fnm{Qingshan} \sur{Xu}}

\author[3]{\fnm{Xiaojuan} \sur{Qi}}

\author[4]{\fnm{Bei} \sur{Yu}}

\author[5]{\fnm{Hanwang} \sur{Zhang}}

\affil[1]{Hefei University of Technology}

\affil[2]{University of Science and Technology of China}

\affil[3]{The University of Hong Kong}

\affil[4]{The Chinese University of Hong Kong}

\affil[5]{Nanyang Technological University}








\abstract{
In this paper, we are the first to examine the correlations between class frequency and the multi-scale noise schedule within diffusion models. For score-based generative models, low-density regions often lead to inaccurately estimated scores, thereby compromising the generation quality. Although the multi-scale noise schedule can alleviate this issue during the diffusion process, low-frequency classes still face the challenge of large low-density regions, resulting in more inaccurate estimated scores than high-frequency classes. Furthermore, high-frequency classes tend to dominate the score space, causing a convergence of most data points towards generating samples from these classes. Consequently, samples generated within low-frequency classes exhibit suboptimal quality and limited diversity. To address this challenge, we propose the \textit{Class-frequency Guided (CFRG)} noise schedule, leveraging the insight that low-frequency classes should be endowed with larger-scale noises. To illustrate the effectiveness of our method, we conduct experiments on various tasks, including image generation, image classification, and text-to-image generation, using imbalanced datasets, \textit{i.e.}, CIFAR-100-LT, and ImageNet-LT. By employing the CFRG noise schedule, we achieve substantial improvements over baselines, manifesting the crucial role of frequency statistics in noise schedule design. 
}




\maketitle

\section{Introduction}
Score-based generative models~\citep{song2020score, ho2020denoising, song2019generative} have garnered considerable attention across diverse domains, spanning image/video generation~\citep{rombach2022high, liu2024sora}, audio synthesis~\citep{chen2020wavegrad}, image editing~\citep{hertz2022prompt, brooks2023instructpix2pix, kawar2023imagic}, and adversarial training~\citep{wang2023better}. Distinguished from GANs or other likelihood-based models, these models concentrate on modeling \textit{the gradient of the log probability density function}, \textit{i.e.}, the score function. A core ingredient in score-based models is the multi-scale noise schedule: data are gradually added with multi-scale Gaussian noise until the signals are diffused into a normal distribution. Subsequently, the model is trained to transit the normal distribution to the data distribution. This process can be theoretically explained by Markov chain~\citep{sohl2015deep, ho2020denoising} or stochastic differential equations (SDEs)~\citep{song2020score}.

The mechanism behind the multi-scale noise schedule attracts lots of attention in the community. Several intriguing properties have been studied in recent work~\citep{song2020score, chen2023importance, hoogeboom2022blurring, kingma2021variational}. Chen et al.~\citep{chen2023importance} reveal the relationship between image size and the noise schedule. P. Kingma et al.~\citep{kingma2021variational} conclude that the diffusion loss is invariant to the shape of the signal-to-noise function SNR(t). Hoogeboom et al.~\citep{hoogeboom2022blurring} explore the Gaussian diffusion process with non-isotropic noise. In this paper, we are the \textit{first} to examine how class frequency relates to the noise schedule in diffusion models.

\begin{figure*}[t]
    \centering
    \subfloat[w/o noise]               { \includegraphics[width=0.23\linewidth]{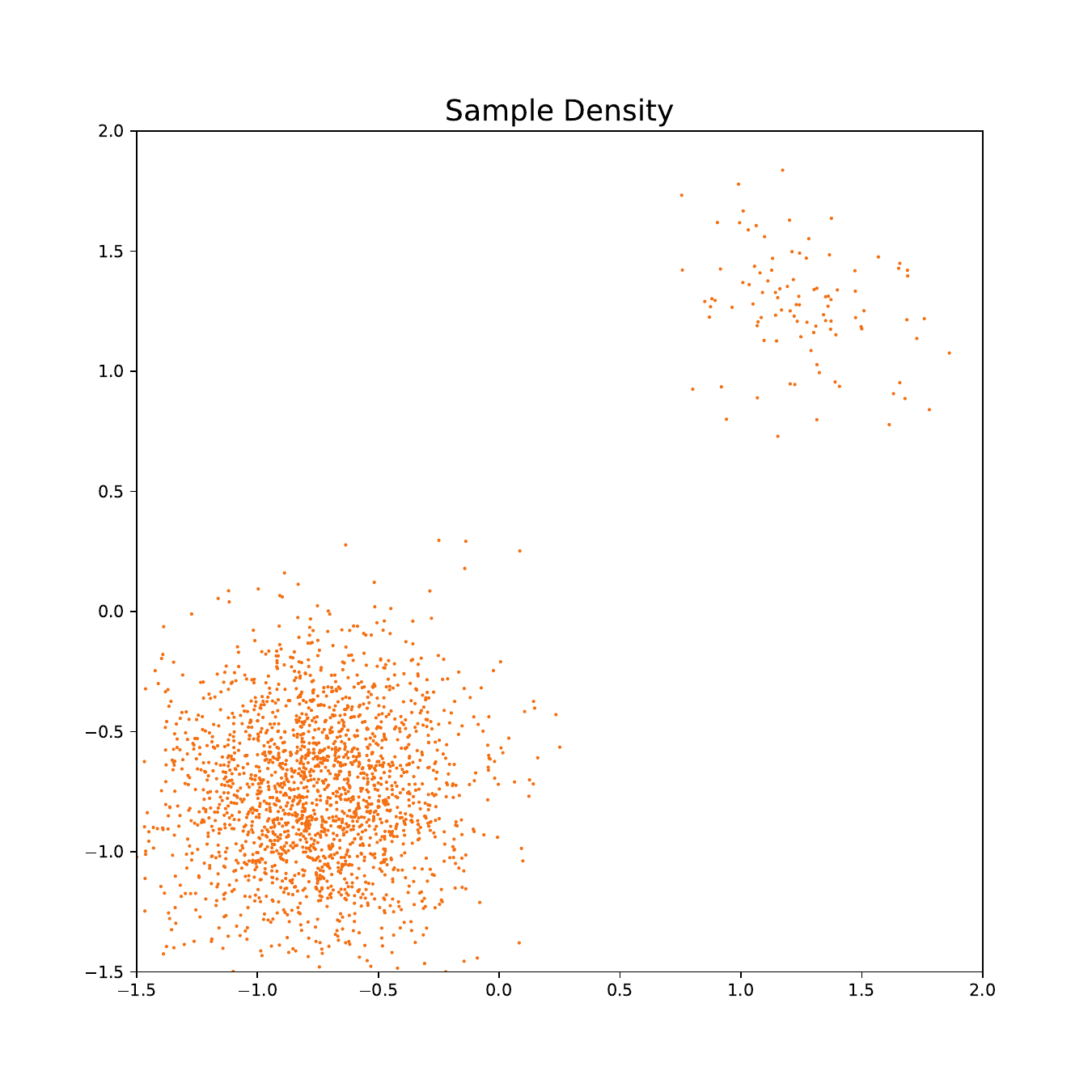} \label{fig:sample_density_clean}}
    \subfloat[w/ Org. noise]  { \includegraphics[width=0.23\linewidth]{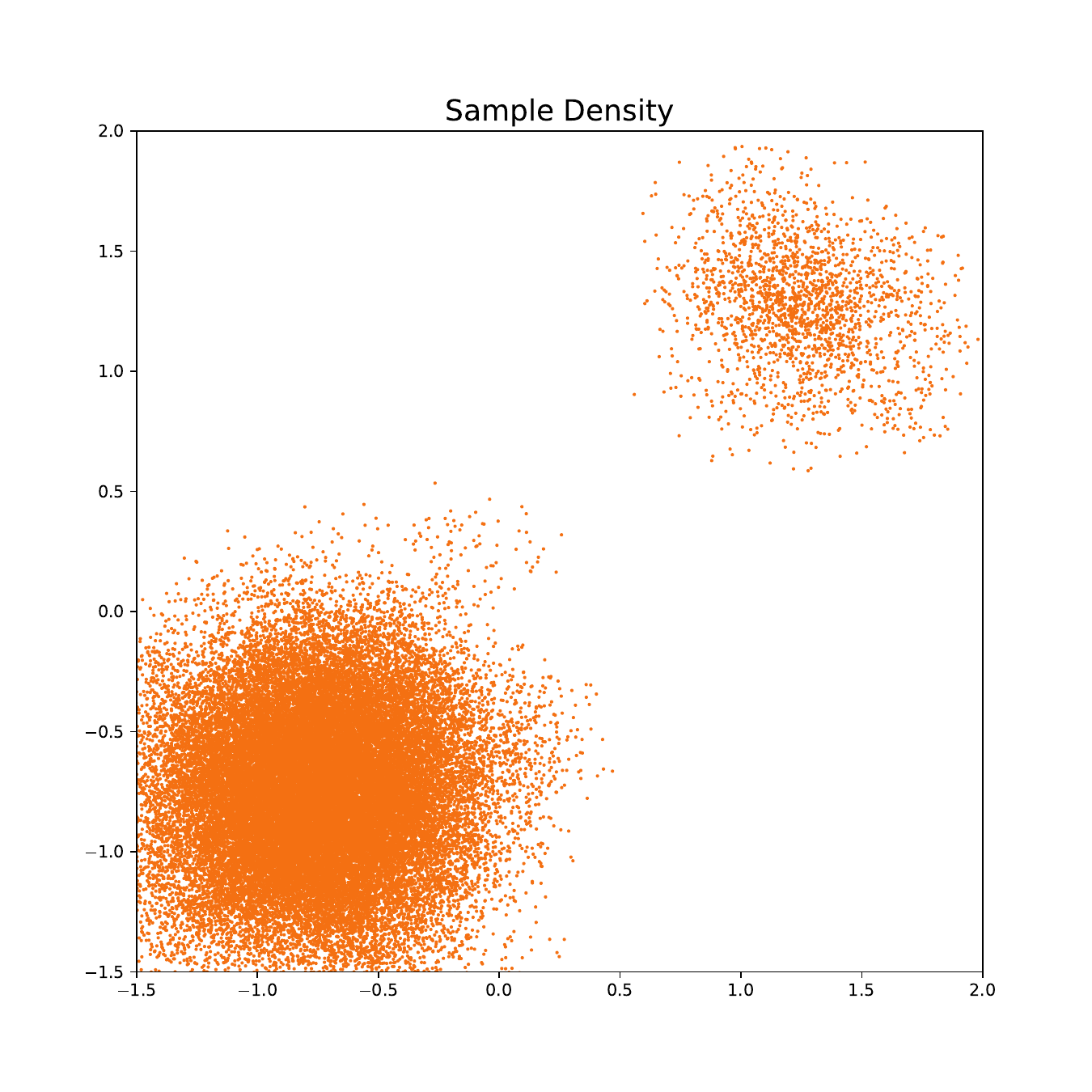}  \label{fig:sample_density_gaussian}}
    \subfloat[w/ CFRG noise]  { \includegraphics[width=0.23\linewidth]{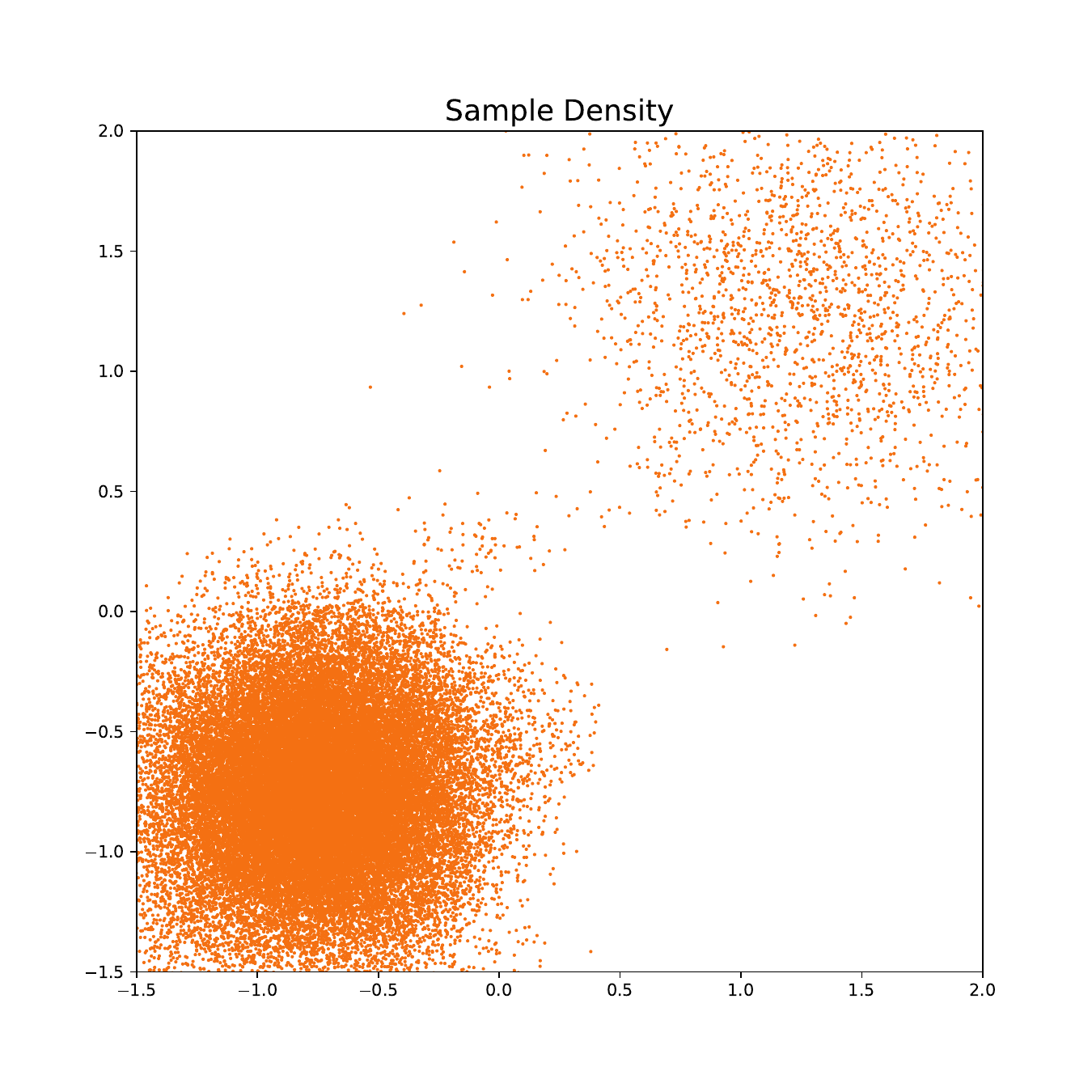}        \label{fig:sample_density_gaussian_balance}}
    \subfloat[CIFAR-100-LT]            { \includegraphics[width=0.28\linewidth]{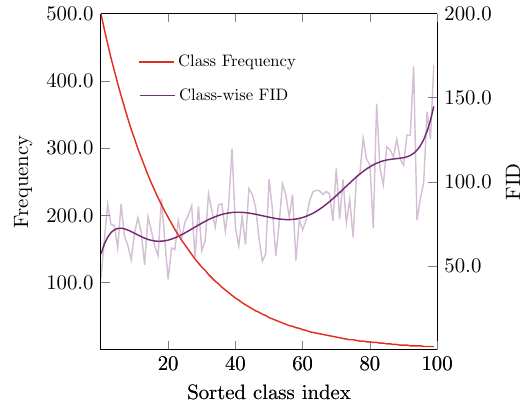}        \label{fig:cifar100lt_freq_fid}}
    \caption{
        \textbf{Sample density visualization.}
        (a) Clean sample density.
        (b) Density of noisy samples perturbed by the equal scale noise for all classes.
        (c) Density of noisy samples perturbed by our class-frequency guided (CFRG) noise schedule.
        Low-frequency classes are more likely to suffer from inaccurate score estimation because of their large low-density regions.
        (d) The quality of generated samples decreases as the class frequency decreases.
    }
    \label{fig:toy_sample_density}
\end{figure*}

\begin{figure*}[h]
    \centering
    \subfloat[Data scores]            { \includegraphics[width=0.248\linewidth]{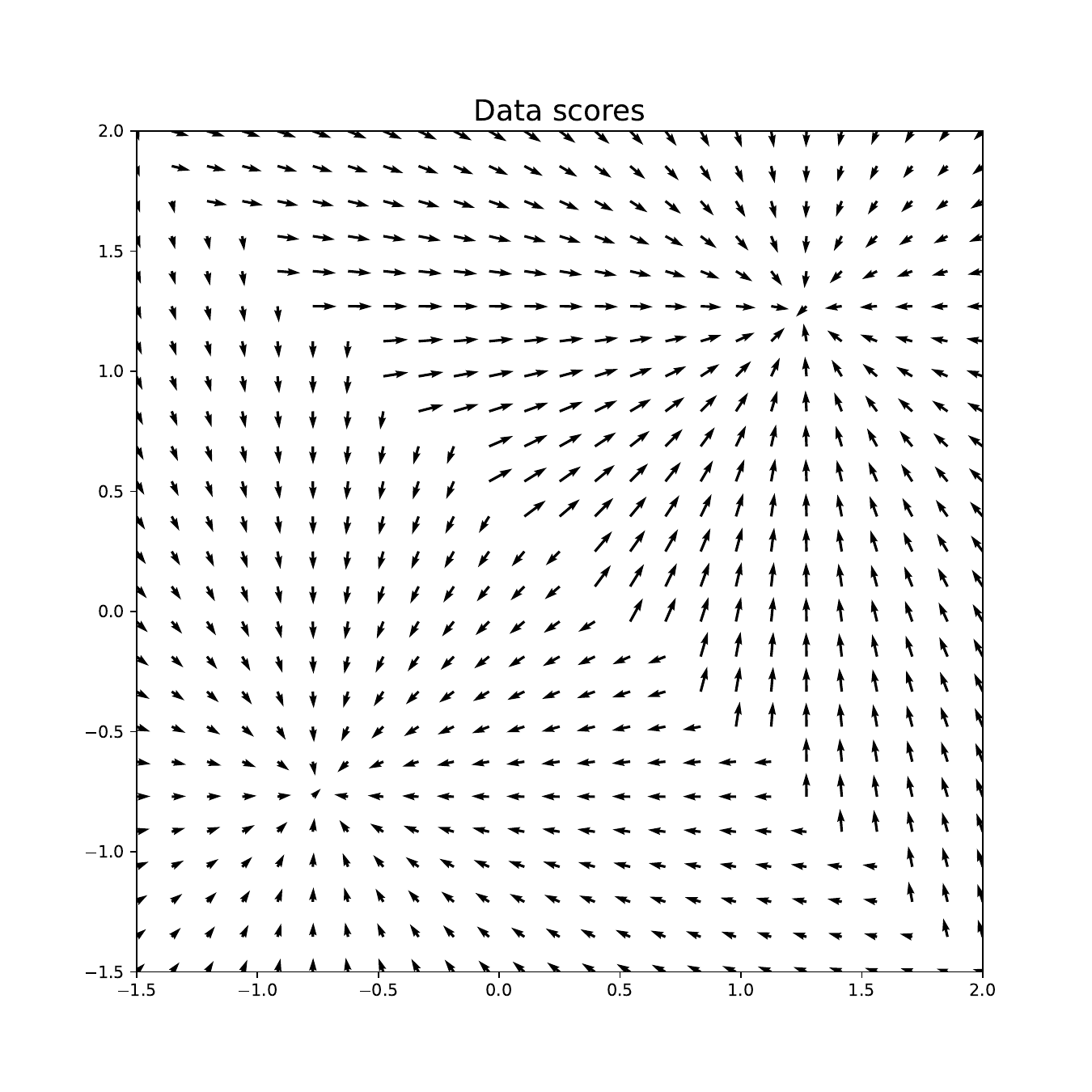}            \label{fig:score_gt}}
    \subfloat[w/o noise]              { \includegraphics[width=0.248\linewidth]{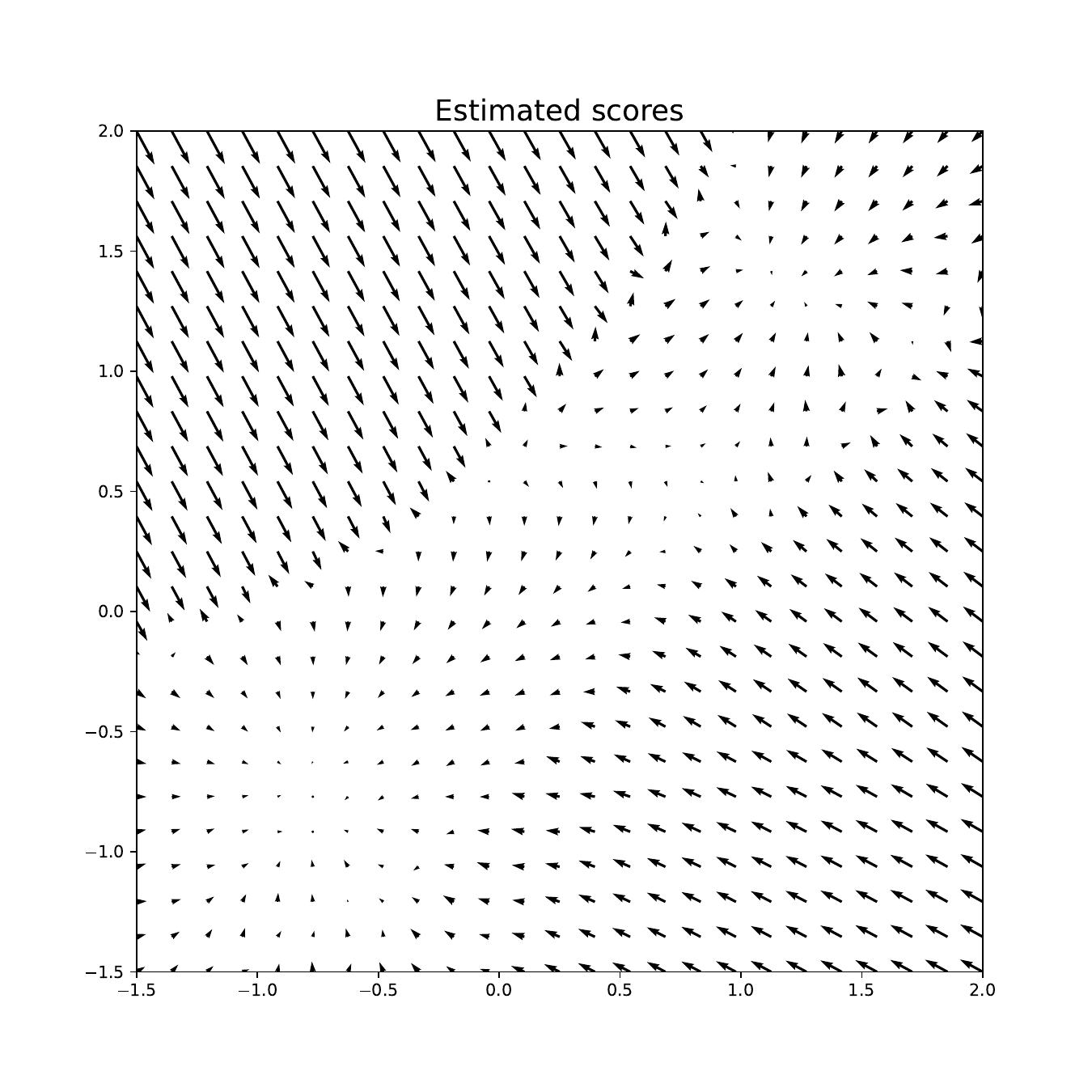}         \label{fig:score_clean}}
    \subfloat[w/ Org.~noise] { \includegraphics[width=0.248\linewidth]{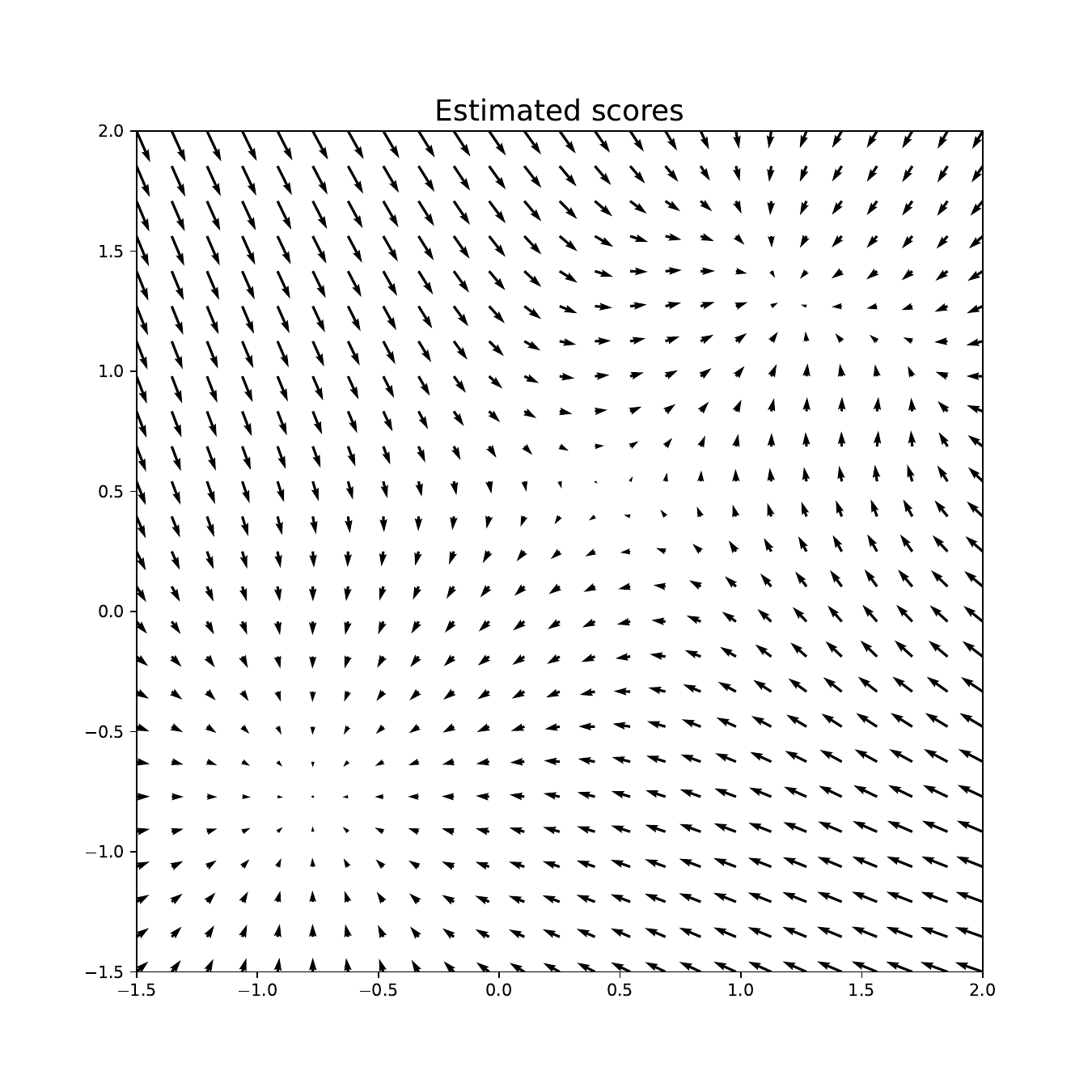}      \label{fig:score_gaussian}}
    \subfloat[w/ CFRG noise] { \includegraphics[width=0.248\linewidth]{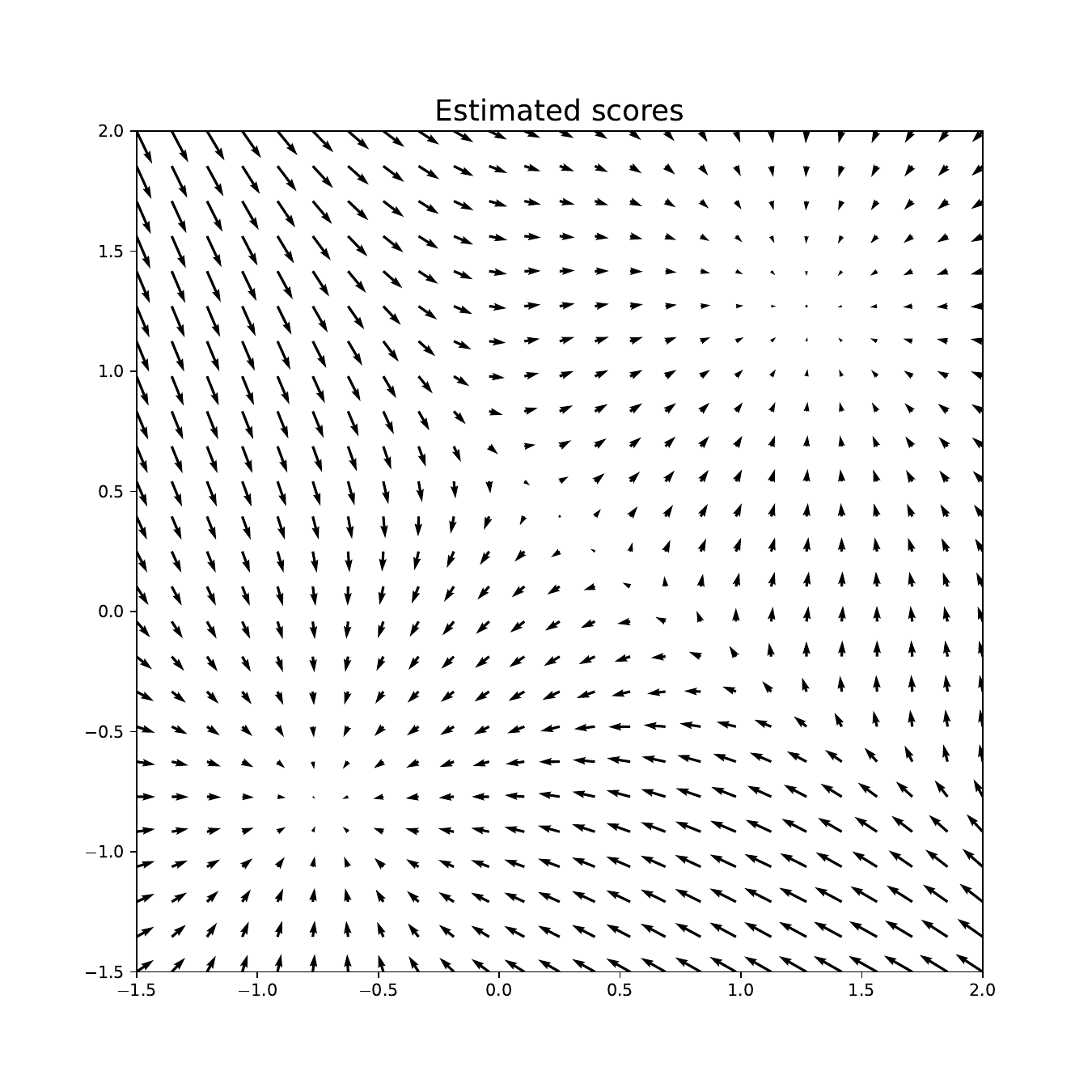} \label{fig:score_gaussian_balance}}
    \caption{
        \textbf{Data score $\nabla_{x} \log p(x)$ space visualization.}
        (a) Ground-truth data scores. 
        (b) Estimated scores with clean samples.
        (c) Estimated scores with samples perturbed by the equal scale noise for all classes.
        (d) Estimated scores with samples perturbed by our class-frequency guided (CFRG) noise schedule. The high-frequency class dominates the estimated score space in (b) and (c). 
        With the CFRG noise schedule, we obtain a more balanced estimated score space for low-frequency classes in (d).
    }
    \label{fig:toy_estimated_scores}
\end{figure*}

\noindent{\bf Our Motivation.}
For score-based generative models such as NCSN~\citep{song2019generative} and DDPM~\citep{ho2020denoising}, the quality of generated samples highly depends on the accuracy of the estimated scores by the learned score function. However, inaccuracies in score estimation can arise, particularly in low-density regions. Despite the integration of multi-scale noise schedules to reduce low-density regions in prominent generative models~\citep{ho2020denoising, song2020score}, \textit{we observe that low-frequency classes continue to face this challenge}.

We present a simplified illustration of a two-Gaussian mixture in Figures~\ref{fig:toy_sample_density} and~\ref{fig:toy_estimated_scores}. In Figure~\ref{fig:sample_density_clean}, the low-frequency class exhibits smaller intra-class variance compared to the high-frequency class due to limited samples. Upon introduction of Gaussian noise, depicted in Figure~\ref{fig:sample_density_gaussian}, the resultant noisy samples populate low-density regions, where $p(x) \approx 0$, for both classes. However, notably larger low-density regions are observed for the low-frequency class, leading to significantly more pronounced inaccuracies in estimated scores.
An additional observation is the dominance of the high-frequency class in the estimated score space. As evidenced in Figures~\ref{fig:score_clean} and~\ref{fig:score_gaussian}, 
most data points converge to generate samples from the high-frequency class,
potentially hindering sample generation for the low-frequency class.
Figure~\ref{fig:cifar100lt_freq_fid} is coherent to these findings: FID scores show a marked increase as class frequency decreases, underscoring substantially lower sample quality in low-frequency classes compared to their high-frequency counterparts.

\noindent{\bf Our Solution.}
To address the above challenges, we propose a \textbf{Class-frequency Guided (CFRG)} noise schedule: \textit{the noise scale should be inversely correlated to class frequency}. 
By applying a relatively larger noise scale to low-frequency classes, we further effectively reduce their low-density regions, as depicted in Figure~\ref{fig:sample_density_gaussian_balance}. 
Additionally, our CFRG noise schedule fosters a more balanced distribution of estimated scores, as evidenced in Figure~\ref{fig:score_gaussian_balance}. 
To assess the effectiveness of our method, we conduct experiments on imbalanced datasets, specifically long-tailed CIFAR~\citep{krizhevsky2009learning} and ImageNet~\citep{imagenet}. In image generation tasks, we achieve FID scores of 5.14 and 2.33 on CIFAR-100-LT and ImageNet-LT, respectively, surpassing the DDPM baseline by \textbf{2.24} and \textbf{0.76}, respectively. Additionally, in image classification, leveraging data generated by our CFRG models yields a notable improvement of \textbf{9.22\%} in top-1 accuracy on CIFAR-100-LT. Finally, we show that our method can be applicable to vision-language diffusion models by text-to-image generation. 
Our key contributions are summarized in what follows:
\begin{itemize}[leftmargin=1.0cm]
    \item We are the \textit{first} to systematically investigate the relationships between the multi-scale noise schedule and class frequency. Two issues on low-frequency classes are identified: larger low-density regions and imbalanced estimated score space.
    \item To solve the challenges, we propose a class-frequency guided (CFRG) noise schedule for diffusion models: the noise scales should be inversely correlated to class frequency.
    \item We validate the effectiveness of our CFRG noise schedule on tasks including image generation, image classification, and text-to-image generation with imbalanced datasets,\textit{i.e.}, CIFAR-100-LT, and ImageNet-LT. 
\end{itemize}

\section{Related Work}

\noindent{\bf Score-based Generative Models.}
Inspired by non-equilibrium statistical physics, Nonequilibrium Thermodynamics (NET~\citep{sohl2015deep}) was the first to deploy a prescribed diffusion process with a Markov chain to gradually transform data into random noise, then reverse the process by training an inverse diffusion model.
The Noise Conditional Score Network (NCSN)~\citep{song2019generative} proposed learning the data distribution by modeling the gradient of the log probability density function, known as the score function. Utilizing multi-scale Gaussian noise, the score function is learned through a score-matching objective, allowing new samples to be generated using annealed Langevin dynamics~\citep{parisi1981correlation} during inference. DDPM~\citep{ho2020denoising} demonstrated for the first time that diffusion models are capable of generating high-quality samples. It also showed the equivalence between diffusion models and denoising score matching across multiple noise levels during training, with annealed Langevin dynamics during sampling. Later, stochastic differential equations (SDEs)~\citep{song2020score} were introduced for score-based models, unifying previous approaches in score-based generative modeling and DDPM.

\noindent{\bf Learning on Imbalanced Data.}
In real-world scenarios, data often follows long-tailed distribution, \textit{i.e.}~a few classes have lots of data while plenty of classes only possess a few samples.
Training on imbalanced data, models exhibit extremely poor accuracy on low-frequency classes.
Re-sampling~\citep{byrd2019effect, buda2018systematic} and Re-weighting~\citep{cui2019class} are two kinds of classical methods to tackle this problem while hurting the representation learning.
Then the classifier and representation learning are decoupled to keep generalizable representations~\citep{kang2019decoupling}.
Methods~\citep{kang2019decoupling, wang2020long, cui2022reslt} have already achieved the best trade-off between high- and low-frequency class performance.
Recently, representation learning techniques~\citep{cui2021parametric, cui2023generalized, Cui_2024_CVPR, cui2024decoupled, 11563882, cui2025generative, zhu2022balanced, du2024probabilistic} have also been developed to address the long-tailed recognition, creating new state-of-the-art performance.
In addition to long-tailed recognition, region rebalance~\citep{cui2022region} and center collapse regularizer~\citep{zhong2023understanding} explore imbalanced learning on semantic segmentation.
Label distribution smoothing (LDS) and feature distribution smoothing (FDS)~\citep{yang2021delving} investigate imbalanced regression.
Class-balancing diffusion models (CBDM)~\citep{qin2023class} expand logits adjustment~\citep{menon2020long} into diffusion models for balanced generation. PoGDiff~\cite{wang2026pogdiff} focuses on text-to-image generation, rebalancing learning regarding the conditional textual feature space density by borrowing statistical strength from neighboring conditions. Unlike existing works, we, in this paper, propose the class frequency guided noise schedule (CFRG) and rebalance image generation learning regarding the noisy image space density by reducing low sample density regions for less frequent classes.

\section{Method}
\subsection{Influences of Class Frequency on Estimated Score $\nabla_{x} \log p(x)$}
\label{sec:influence_cf}

\noindent{\bf Sampling with Score Function.}
Score-based models~\citep{ho2020denoising, song2019generative, song2020score} learn the data distribution's probability density via a score function, \textit{i.e.}, \textit{the gradient of the log probability density function $\nabla_{x} \log p(x)$}. Thanks to the Langevin dynamics~\citep{parisi1981correlation}, new samples could be generated with the learned score function in an iterative manner as follows:
\begin{equation}
    x_{i+1} = x_{i} + \eta \nabla_{x} \log p(x) + \sqrt{2\eta} \epsilon, i=0,1,...,K,
    \label{eq:langevin}
\end{equation}
where $x_{0} \sim \pi(x)$ is a prior distribution, $\epsilon \sim \mathcal{N}(\mathbf{0}, \mathbf{I})$, $\eta \to 0$ is the step size, K $\to \infty$ is the number of steps for new sample generation. NCSN~\citep{song2019generative} extends the sampling process to annealed Langevin dynamics with a multi-scale noise schedule. 

Markov chain in DDPM~\citep{ho2020denoising} is adopted to sampling with the score function at inference:
\begin{equation}
    x_{t-1} = \frac{1}{\sqrt{1-\sigma_{t}}}(x_{t} + \sigma_{t} \nabla_{x_{t}} \log p(x_{t})) + \beta_{t} \epsilon,
    \label{eq:markov}
\end{equation}
where $0<\sigma_{1}<\sigma_{2}<...<\sigma_{T}<1$ is the multi-scale noise schedule in the diffusion process, the maximum time-step T=1000, $x_{T} \sim \mathcal{N}(\mathbf{0}, \mathbf{I})$ is a prior distribution, $\epsilon \sim \mathcal{N}(\mathbf{0}, \mathbf{I})$, $\beta_{t}$ is a function of $\sigma_{1:t}$. The derivation for Eq.~\eqref{eq:markov} is shown in Appendix~\ref{sec:score_function_ddpm}.

Additionally, diffusion and reverse diffusion processes can be equivalently represented with forward and reverse stochastic differential equations (SDEs)~\citep{song2020score}. Specifically, sampling with the reverse SDE at inference is defined as:
\begin{equation}
    dx = [f(x,t)-g(t)^2 \nabla_{x} \log p_{t}(x)]dt + g(t)d\overline{w},
    \label{eq:sde}
\end{equation}
where $\{ x(t) \}_{t=0}^{T}$ is a diffusion process with a continuous variable $t \in [0, T]$, $f(\cdot, t): \mathbb{R}^{d}->\mathbb{R}^{d}$ is a vector-valued function called the drift coefficient of $x(t)$, $g(t)$ is a scalar function known as the diffusion coefficient of $x(t)$, $p_{t}(x)$ is the probability density of $x(t)$, $\overline{w}$ is a standard Wiener process.

\noindent{\bf Low-density Regions Lead to Inaccurate Score Estimation.}
\textit{The quality of generated samples with Eq.~\eqref{eq:langevin}, Eq.~\eqref{eq:markov}, and Eq.~\eqref{eq:sde} heavily relies on the learned score function $\nabla_{x} \log p(x)$}. However, the estimated scores often prove inaccurate, particularly in the initial sampling stages. This phenomenon arises due to low sample density regions of $p(x)$, leading to insufficient training under the following objective: 
\begin{equation}
    \mathbb{E}_{p(x)} ||\nabla_{x} \log p(x) - s_{\theta}(x)||_{2}^{2},
    \label{eq:score_matching}
\end{equation}
where $s_{\theta}$ is a neural network for score estimation. Due to the unavailability of the real data score, it is often implemented with score-matching techniques. The loss function in DDPM~\citep{ho2020denoising} is also equivalent to Eq.~\eqref{eq:score_matching}, which is theoretically evidenced in Appendix~\ref{sec:inaccurate_score_ddpm}.

To reduce the low-density regions, multiple scales of noise perturbations are adopted in recent diffusion models~\citep{song2019generative, ho2020denoising, song2020score}. With an increasing noise schedule $0 < \sigma_{1} < \sigma_{2} <...<\sigma_{T} <1$, an clean image $x_{0}$ is diffused into $x_{T} \in \mathcal{N}(\mathbf{0}, \mathbf{I})$ when $T \rightarrow \infty$, \textit{i.e.}, 
$x_{t} = \sqrt{1-\sigma_{t}} x_{t-1} + \sigma_{t} \epsilon$, 
where $\epsilon \in \mathcal{N}(\mathbf{0},\mathbf{I})$. Then the score function is trained on noisy samples with each scale of $\sigma_{t}$:
\begin{equation}
    \mathbb{E}_{p_{\sigma_{t}}(x_{t})} ||\nabla_{x_{t}} \log p_{\sigma_{t}}(x_{t}) - s_{\theta}(x_{t}, t)||_{2}^{2}.
\end{equation}

Figures~\ref{fig:toy_sample_density} and~\ref{fig:toy_estimated_scores} show a toy example of a mixture of two Gaussians.
As shown in Figures~\ref{fig:sample_density_clean} and~\ref{fig:sample_density_gaussian}, the low-density regions are reduced after adding Gaussian noise to clean examples.
Correspondingly, the estimated scores become much more accurate as illustrated by Figures~\ref{fig:score_gt},~\ref{fig:score_clean}, and~\ref{fig:score_gaussian}. 

\noindent{\bf The Effects of Class Frequency on Score-based Generative Models.}
The multi-scale noise schedule significantly contributes to accurate score estimation, essential for generating high-quality samples. In this paper, we investigate the influence of class frequency on score-based generative models and establish that \textit{class frequency is also a crucial factor in multi-scale noise schedule design}. Our analysis reveals two primary observations:
\begin{itemize}[leftmargin=1.0cm]
    \item Equipped with the original multi-scale noise schedule (all classes are equally treated), low-frequency classes still encounter problems of large low-density regions. Consequently, their estimated scores $\nabla_{x} \log p(x)$ tend to be more inaccurate compared to high-frequency classes.
    \item Training on imbalanced data, the estimated score $\nabla_{x} \log p(x)$ space is dominated by high-frequency classes, impairing the generation quality of low-frequency classes.
\end{itemize}

In the toy example depicted in Figures~\ref{fig:toy_sample_density} and~\ref{fig:toy_estimated_scores}, benefiting from abundant samples, the high-frequency class enjoys a larger intra-class variance, whereas the low-frequency class displays a smaller variance (see Figure~\ref{fig:sample_density_clean}).
With the Gaussian noise (see Figure~\ref{fig:sample_density_gaussian}), low-density regions where $p(x) \approx 0$ are significantly reduced for both high- and low-frequency classes. However, high-density regions for the low-frequency class remain notably smaller than those for the high-frequency class, suggesting potentially greater inaccuracies of estimated scores for the former.
Additionally, as shown in Figures~\ref{fig:score_clean} and~\ref{fig:score_gaussian}, the estimated scores for the majority of data points converge towards generating samples from the high-frequency class. This dominance within the estimated score space poses a significant challenge for generating high-quality samples from the low-frequency class.

\noindent{\bf A Case Study.}
With the above analysis, we conclude that class frequency is a crucial factor in the noise schedule design for score function learning. Further, we confirm that \textit{the noise scale should be inversely correlated to class frequency} with a case study of DDPM~\citep{ho2020denoising}. DDPM~\citep{ho2020denoising} uses a fixed linear noisy schedule,
\begin{equation}
    \sigma_{t} = (\sigma_{T} - \sigma_{1}) \frac{t-1}{T-1} + \sigma_{1}.
    \label{eq:ddpm_noisy_schedule}
\end{equation}
With the default hyper-parameter $\sigma_{1}=1e-4$, we examine the effects of various values of $\sigma_{T}$ on the quality of generated samples between classes with different class frequencies. Experiments are conducted on CIFAR-100-LT and ImageNet-LT datasets. The imbalanced factors ($\frac{N_{max}}{N_{min}}$, where $N$ is the number of samples in the class) are 100 and 256 respectively. Classes are grouped into ``Many", ``Medium" and ``Few" according to class frequencies.
We report the per-group and overall FID for evaluation. Especially, the per-group FID is calculated by averaging the FID of all classes in the group. The experimental results are summarized in Table~\ref{tab:exp_ablation}.

Results in Table~\ref{tab:exp_ablation} reveal two interesting phenomenon:
\begin{itemize}[leftmargin=0.6cm]
    \item A proper $\sigma_{T}$ reduces low-density regions, contributing to accurate score estimation. On the other hand, a too large $\sigma_{T}$ can over-corrupt the data, leading to hard optimization of the reverse diffusion process. Thus, $\sigma_{T}$ should be carefully chosen to achieve a good overall FID score.
    \item The noise scale $\sigma_{T}$ should be inversely correlated to class frequency. For low-frequency classes, a relatively higher $\sigma_{T}$ can enlarge their high-density regions and thus potentially raise the accuracy of estimated scores. 
\end{itemize}

\begin{table}[tb!]
    \centering
    \caption{
        \textbf{Larger noise scale benefits low-frequency classes.}
        Exploration of noise schedule effects regarding the class frequency on CIFAR-100 with an imbalance factor of 100, and ImageNet-LT with an imbalance factor of 256.}
    \setlength{\tabcolsep}{0.03in}
    {
        \begin{tabular}{lccccc}
            \toprule
            Dataset &sigma &Many &Medium &Few &Overall\\
            \midrule
            \multirow{4}{*}{CIFAR-100LT} &0.02    &\textbf{68.72} &\textbf{80.61} &108.44 &7.38 \\
            & 0.05    &70.60 &81.93 &\textbf{106.33} & 7.56\\
            & 0.07    &72.63 &83.64 &107.81 &8.20 \\
            & 0.10    &74.41 &86.71 &109.75 &9.13 \\
            \midrule
            \multirow{3}{*}{ImageNetLT} &0.01  &\textbf{156.98} &167.27 &177.59 &3.09 \\
            & 0.02  &161.71 &167.24 &\textbf{176.54} &3.18 \\
            & 0.03  &158.50 &168.66 &178.57 &3.75 \\
            \bottomrule
        \end{tabular}%
    }
    \label{tab:exp_ablation}
    \vspace{-0.1in}
\end{table}

\subsection{Class-frequency Guided Noise Schedule}
The analysis in Sec~\ref{sec:influence_cf} highlights the importance of class frequency in multi-scale noise schedule design for score function learning and reveals that the noise scale should be inversely correlated to class frequency. Based on this insight, we are the \textit{first} to consider the class frequency for the multi-scale noise schedule designing and propose a \textbf{Class-frequency Guided (CFRG)} noise schedule for score-based generative models.

Compared to high-frequency classes, low-frequency classes are adversely affected by large low-density regions, leading to inaccurate score estimation.
In the toy example of Figure~\ref{fig:toy_sample_density} and Figure~\ref{fig:toy_estimated_scores},
we consider the high-frequency class $A \sim \mathcal{N}(\mathbf{\mu_{A}}, \mathbf{\sigma_{A}})$,
the low-frequency class $B \sim \mathcal{N}(\mathbf{\mu_{B}}, \mathbf{\sigma_{B}})$, and $\mathbf{\sigma_{A}} \gg \mathbf{\sigma_{B}}$.
With the fixed linear noisy schedule in Eq.~\eqref{eq:ddpm_noisy_schedule},
$x_{t}=\sqrt{\bar \alpha_{t}}x_{0} + \sqrt{1-\bar \alpha_{t}} \epsilon$, where $\bar \alpha_{t}=\Pi_{i=1}^{t} \alpha_{i}$, $\alpha_{t}=1-\sigma_{t}$, $\epsilon \in \mathcal{N}(\mathbf{0}, \mathbf{I})$. Then, the following equations are established for classes $A$ and $B$ respectively:
\begin{eqnarray}
     x_{t} &\sim& \mathcal{N}(\sqrt{\bar \alpha_{t}} \mathbf{\mu_{A}}, \sqrt{-\bar \alpha_{t}(1-\mathbf{\sigma_{A}}^2)+1}), \label{eq:class_a_xt} \\
     x_{t} &\sim& \mathcal{N}(\sqrt{\bar \alpha_{t}} \mathbf{\mu_{B}}, \sqrt{-\bar \alpha_{t}(1-\mathbf{\sigma_{B}}^2)+1}),
     \label{eq:class_b_xt}
\end{eqnarray}
where $\mathbf{\sigma_{A}}<\mathbf{1}$ and $\mathbf{\sigma_{B}}<\mathbf{1}$, Normalization operations are used to pixel inputs for the constraint.

Since $\mathbf{\sigma_{A}} \gg \mathbf{\sigma_{B}}$, we know that $\sqrt{-\bar \alpha_{t}(1-\mathbf{\sigma_{A}}^2)+1} > \sqrt{-\bar \alpha_{t}(1-\mathbf{\sigma_{B}}^2)+1}$ for all of time-step $t$, which means noisy samples of class $B$ always have smaller high-density regions than class $A$ and thus there are larger low-density regions for class $B$. 

$\mathbf{\sigma_{A}}$ and $\mathbf{\sigma_{B}}$ are constant determined by the training data. We thus schedule a class-wise $\bar \alpha_{t}$ and propose the \textbf{Class-frequency Guided (CFRG)} noise schedule to enlarge the high-density regions of low-frequency classes. In detail, for a specific class $i$,
\begin{eqnarray}
    \sigma_{T}^{i} \!&=& \frac{\sigma_{T}^{max} \!-\! \sigma_{T}^{min}}{F_{min} \!-\! F_{max}} (F_{i}-F_{max}) \!+\! \sigma_{T}^{min}, \label{eq:sigma_T}\\
    \sigma_{t}^{i} \!&=& (\sigma_{T}^{i} \!-\! \sigma_{1}) \frac{t-1}{T-1} \!+\! \sigma_{1}, \label{eq:sigma_t}   
\end{eqnarray}
where $F_{min}$ and $F_{max}$ are the least and most class frequency respectively, $F_{i}$ is the frequency of class $i$, $\sigma_{T}^{min}$ and $\sigma_{T}^{max}$ are tunable hyper-parameters, $\sigma_{1}$ is set to 1e-4 following DDPM~\citep{ho2020denoising}. We also explore the effective number~\citep{cui2019class} for the calculation of $F$ values.

With the constraints $F_{A} > F_{B} > F_{C}$ for any 3 classes $A \sim \mathcal{N}(\mathbf{\mu_{A}}, \mathbf{\sigma_{A}})$, $B \sim \mathcal{N}(\mathbf{\mu_{B}}, \mathbf{\sigma_{B}})$, and $C \sim \mathcal{N}(\mathbf{\mu_{C}}, \mathbf{\sigma_{C}})$ in the dataset, Eq.~\eqref{eq:sigma_T} and Eq.~\eqref{eq:sigma_t} guarantee that:
\begin{eqnarray}
    -\bar \alpha_{t}^{A}&=&-\Pi_{i=1}^{t} (1-\sigma_{i}^{A}) < -\bar \alpha_{t}^{B}, \\
    -\bar \alpha_{t}^{B}&=&-\Pi_{i=1}^{t} (1-\sigma_{i}^{B}) <-\bar \alpha_{t}^{C}, \\
    -\bar \alpha_{t}^{C}&=&-\Pi_{i=1}^{t} (1-\sigma_{i}^{C}) <0,
\end{eqnarray}
which implies that the less class frequency, the more compensation in terms of high-density regions will be provided, thus benefiting the generation quality of low-frequency classes.

\subsection{Analysis}
\noindent{\bf Density Analysis.} Eqs.~\eqref{eq:langevin},~\eqref{eq:markov} and~\eqref{eq:sde} show that the quality of generation heavily depends on the accuracy of the learned score function. Furthermore, Eq.~\eqref{eq:score_matching} suggests that low sample density --- particularly in low-frequency classes --- can lead to undertrained score estimates, ultimately degrading generation quality.

In Table~\ref{tab:density_error}, we quantify the low-density regions ($\leq \delta$) of imbalanced data under different noise schedules, highlighting that the CFRG noise schedule can improve model generation quality by reducing low-density regions. Please refer to Algorithm~\ref{algorithm:density_analysis} in the Appendix for more details.

\begin{table}[t]
    \centering
    \caption{\textbf{Density analysis}. Quantity of low-density regions for DDPM and CFRG noise schedule. Please refer to Algorithm~\ref{algorithm:density_analysis} in the supplementary file for more details.}
    {
    \begin{tabular}{cccc}
    \toprule
         Method & $\delta$=-146.19 & $\delta$=-143.90 &$\delta$=-142.33 \\
         \midrule
         DDPM(baseline) &14.95\% &25.51\% &35.15\% \\
         CFRG &\textbf{10\%} &\textbf{20\%} &\textbf{30\%} \\
         \bottomrule
    \end{tabular}
    }
    \label{tab:density_error}
    \vspace{-0.1in}
\end{table}

\noindent{\bf Knowledge Transfer from High-frequency Classes to Low-frequency Ones.}
Training with a single low-frequency class suffers from extremely limited data, while training the entire data together enables knowledge transfer from high-frequency classes to low-frequency classes for common patterns in images. 

\begin{table}[t]
    \centering
    \setlength{\tabcolsep}{0.04cm}
    \caption{\textbf{Knowledge transfer}. Training with high- and low-frequency classes together can benefit the generation quality of low-frequency classes.}
    \vspace{-0.1in}
    \begin{tabular}{ccc}
    \toprule
     Method & FID-All($\downarrow$) & FID-50($\downarrow$) \\
     \midrule
     DDPM(50 least-frequent classes)  &47.96  &27.78 \\
     DDPM(all data)    &7.38  &13.25 \\
     \midrule
     CFRG(all data)    &\textbf{6.62} &\textbf{12.19} \\
     \bottomrule
    \end{tabular}
    \vspace{-0.1in}
    \label{tab:knowledge_transfer}
\end{table}

To illustrate the knowledge transfer between low-frequency and high-frequency classes, we train a model with data of 50 least-frequent classes as a baseline on CIFAR-100-LT. We report \textit{FID-All} (FID on all classes) and \textit{FID-50}(FID on the 50 least-frequent classes) respectively. Table~\ref{tab:knowledge_transfer} shows the empirical results. 
We observe that the DDPM model trained on the whole data achieves much lower FID on the 50 least-frequent classes when compared with the model that traiend only with the data of 50 least-frequent classes, indicating the knowledge transfer from high-frequency classes to low-frequency classes.

\section{Experiments}
\label{sec:experiments}
In Section~\ref{sec:ablation}, we conduct ablations on the effects of hyper-parameters and designs in our CFRG noise schedule. Comparisons in image generation are presented in Section~\ref{sec:generation}. We also discuss how generated samples benefit image classification tasks in Section~\ref{sec:image_classification}. The potential of our CFRG method on vision-language diffusion models is confirmed in Appendix~\ref{sec:text_to_image}. For more details on experimental settings, please refer to Appendix~\ref{sec:exp_config}.

\subsection{Ablation Experiments}
\label{sec:ablation}
\noindent{\bf Comparison with Re-sampling and Re-weighting}.
Long-tailed learning has been widely researched in classification and regression. However, these works are hard to or even can't transfer to image generation tasks. We have also included resampling-based methods for comparisons in Table~\ref{tab:sota_cifar100lt_imagenetlt}. With class-balanced resampling (RS) and SQRT resampling, the generation model even achieves worse performance than DDPM baseline. This observation is consistent with findings in previous work~\citep{qin2023class}. Here, we include empirical results of the re-weighting method in Table~\ref{tab:re_weight}. 
 \begin{table}[h]
     \centering
     \setlength{\tabcolsep}{0.04in}
     \caption{\textbf{Comparisons with re-sample and re-weight methods on CIFAR-100-LT.}}
     \begin{tabular}{cccccc}
          \toprule
          Method &  DDPM &RS &SQRT-RS &Re-weight &\textbf{CFRG} \\
          \midrule
          FID($\downarrow$)    & 7.38 &10.50 &9.72 &7.18 &\textbf{6.62} \\
          \bottomrule
     \end{tabular}
     \label{tab:re_weight}
 \end{table}

\noindent{\bf Ablation on Form of CFRG Noise Schedule.}
Class-frequency guided (CFRG) noise schedule in Eqs.~\eqref{eq:sigma_T} and~\eqref{eq:sigma_t} assigns $\sigma_{T}^{i}$ for class $i$ with a linear function in terms of its class frequency $F_{i}$. Here we conduct ablations on another possible form: 
\begin{equation}
     \sigma_{T}^{i} = \frac{\sigma_{T}^{\max} - \sigma_{T}^{\min}}{C-1} (i-1) + \sigma_{T}^{\min}, \label{eq:sigma_T_linear}\\
\end{equation}
where classes are sorted by their frequencies and class $i$ has the $i$th maximum frequency, $C$ is the number of classes.
Eq.~\eqref{eq:sigma_T_linear} only consider the rank of class frequency rather than its actual value.

Additionally, considering the similarities among samples, the effective number of frequency for class $i$ can be calculated with the following equation~\citep{cui2019class}:
\begin{equation}
    F_{i}=\frac{1 - \gamma^{N^{i}}}{1-\gamma}, \gamma \in (0,1),
    \label{eq:effective_num}
\end{equation}
where $N^{i}$ is the number of samples in class $i$.

\begin{figure*}[t]
    \centering
    \subfloat[CFRG noise schedule]            { \includegraphics[width=0.38\linewidth]{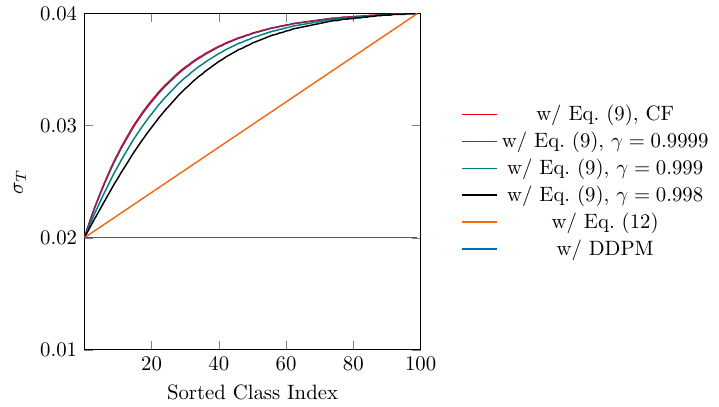} \label{fig:sigma_T_cfg}}
    \subfloat[Guidance scale $\omega$]            { \includegraphics[width=0.36\linewidth]{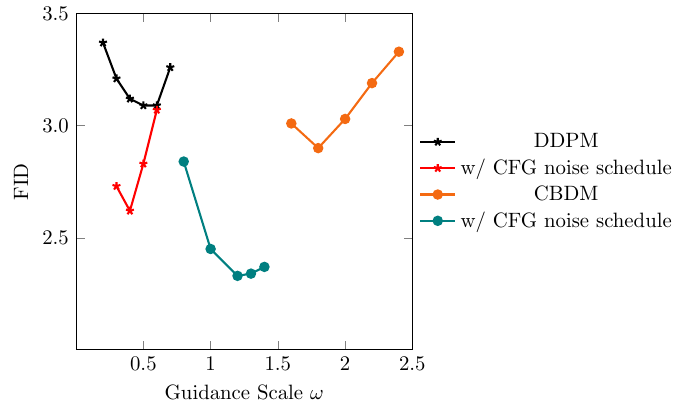}  \label{fig:guidance_scale}}
    \subfloat[CIFAR-100-LT]{ \includegraphics[width=0.236\linewidth]{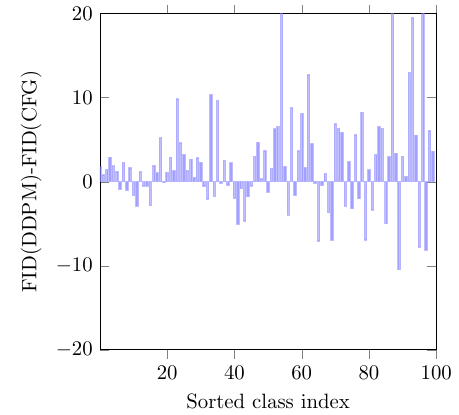}        \label{fig:cifar100_diff_fid}}
    \vspace{-0.1in}
    \caption{
        \textbf{Ablation studies.}
        (a) The curves of $\sigma_{T}$ with different form of CFRG noise schedule.
        (b) The effects of guidance scale $\omega$ with classifier free guidance on ImageNet-LT. The image size $32 \times 32$ is used.
        (c) The CFRG noise schedule significantly benefits low-frequency class performance.  
    }
    \label{fig:ablation}
    \vspace{-0.1in}
\end{figure*}

The function curves for $\sigma_{T}^{i}$ from Eqs.~\eqref{eq:sigma_T_linear} and~\eqref{eq:sigma_T} with various values of $\gamma$ are shown in Figure~\ref{fig:sigma_T_cfg}. As $\gamma$ approaches 1.0, the effective number of frequencies becomes close to the original class frequencies. 
On CIFAR-100-LT, with a proper $\gamma=0.999$, the effective imbalance factor is reduced from 100 to 78.88 meanwhile representing the intra-class variance well and thus achieving good performance.  
We summarize experimental results on CIFAR-100-LT in Table~\ref{tab:ablation_cfg}.
Compared to Eq.~\eqref{eq:sigma_T}, Eq.~\eqref{eq:sigma_T_linear} assigns the $\sigma_{T}^{i}$ for class $i$ only considering the rank information of its class frequency. However, the rank can not accurately reflect the intra-class variance.
As shown in Table~\ref{tab:ablation_cfg}, with Eq.~\eqref{eq:sigma_T} and a $\gamma=0.999$,
our model achieves 6.62 FID, outperforming w/ Eq.~\eqref{eq:sigma_T_linear} by 0.68 and thus demonstrating the importance of the frequency statistics. Meanwhile, our model surpasses the DDPM baseline by \textbf{0.76} FID, showing the effectiveness of our CFRG noise schedule.

\begin{table*}[t]
\centering%
\begin{minipage}{0.55\linewidth}
\centering%
\caption{\textbf{CFRG noise schedule designing.} }
\begin{tabular}{ccc}
    \toprule
    Settings &EIF &FID \\
    \midrule
    w/ baseline (DDPM) &100 &7.38 \\
    \midrule
    w/ Eq.~\eqref{eq:sigma_T}, F&100 &6.74 \\
    w/ Eq.~\eqref{eq:sigma_T}, $\gamma=0.9999$&97.56 &6.74 \\
    w/ Eq.~\eqref{eq:sigma_T}, $\gamma=0.999$ &78.88 &6.62 \\
    w/ Eq.~\eqref{eq:sigma_T}, $\gamma=0.998$ &63.50 &7.10 \\
    w/ Eq.~\eqref{eq:sigma_T_linear} &- &7.30 \\
   \bottomrule
   \end{tabular}
   \label{tab:ablation_cfg}
\end{minipage}%
\hspace{0.03in}
\begin{minipage}{0.43\linewidth}
\centering%
\caption{\textbf{Ablation study of $\sigma_{T}^{\max}$.}}
\begin{tabular}{ccc}
\toprule
Dataset &$\sigma_{T}^{\max}$ &FID \\
\midrule
CIFAR-100-LT &0.03 &6.98 \\
CIFAR-100-LT &0.04 &6.62 \\
CIFAR-100-LT &0.05 &6.86 \\
\midrule
ImageNet-LT  &0.02  &2.62 \\
ImageNet-LT  &0.025 &2.84 \\
ImageNet-LT  &0.03  &2.90 \\
\bottomrule
\end{tabular}
\label{tab:sigma_T_max}
\end{minipage}
\vspace{-0.1in}
\end{table*}

\begin{table*}[t]
    \centering
    \caption{
        \textbf{Improvement analysis on ImageNet-LT.}
        We report FIDs on ``Many'', ``Medium'', ``Few'', and Overall classes. For ``Many", ``Medium'' and ``Few'' groups, we calculate class-wise FIDs and average the FIDs of classes in the group.  
    }
    \setlength{\tabcolsep}{0.25in}
    \label{tab:improvement_imagenetlt}
    {
        \begin{tabular}{lcccc}
            \toprule
            Method &Many &Medium &Few &Overall\\
            \midrule
            Baseline (DDPM)  &\textbf{156.98} &167.27 &177.59 &3.09 \\
            \midrule
            \textbf{CFRG(Ours)} &158.39 &165.81(\textbf{-1.64}) &\textbf{174.57(-3.02)} &2.62 \\
            +CBDM &158.21 &\textbf{165.29(-1.98)} &175.83(\textbf{-1.76}) &2.33 \\
            \bottomrule
        \end{tabular}
    }
\end{table*}

\noindent{\bf Ablation on $\sigma_{T}^{\min}$ and $\sigma_{T}^{\max}$ in CFRG Noise Schedule.}
Following DDPM~\citep{ho2020denoising}, we apply $\sigma_{1}=1e-4$ to our all experiments.
As shown in Table~\ref{tab:exp_ablation}, $\sigma_{T}=0.02$ and $\sigma_{T}=0.01$ are the best choices for DDPM on CIFAR-100-LT and ImageNet-LT separately.
In CFRG noise schedule, we also use $\sigma_{T}^{\min}=0.02$ and $\sigma_{T}^{\min}=0.01$ for CIFAR-100-LT and ImageNet-LT respectively.
The ablation for $\sigma_{T}^{\max}$ is presented in Table~\ref{tab:sigma_T_max}.
A relatively larger $\sigma_{t}^{i}$ can reduce the low-density regions as analyzed in Section~\ref{sec:influence_cf}, leading to more accurate estimated scores. However, a too large $\sigma_{t}^{i}$ can also over-corrupt data and alter it significantly from the original distribution, resulting in hard optimization of the reverse diffusion process. Thus, a reasonable $\sigma_{T}^{\max}$ should be adopted.
Table~\ref{tab:sigma_T_max} illustrates that $\sigma_{T}^{\max}=0.04$ and $\sigma_{T}^{\max}=0.02$ are the best choices on CIFAR-100-LT and ImageNet-LT individually.

\noindent{\bf Improvements on Low-frequency Classes.}
Our findings show that the noise scale should be inversely related to class frequency in the multi-scale noise schedule of diffusion models. Empirically, models trained with our class-frequency guided (CFRG) noise schedule achieve significant overall improvements regarding both sample quality and diversity. Here, we illustrate that the CFRG noise schedule benefits the low-frequency classes. With experiments on CIFAR-100-LT, we measure the difference in class FIDs between the DDPM model and our CFRG model. As shown in Figure~\ref{fig:cifar100_diff_fid}, our model achieves much lower FIDs on low-frequency classes, manifesting higher generation quality benefiting from our proposed CFRG noise schedule.
Besides, we analyze the improvements on the ImageNet-LT dataset.
Table~\ref{tab:improvement_imagenetlt} summarizes our experimental results. After applying our CFRG noise schedule, the ``Medium'' and ``Few'' classes achieve much better generation performance when compared to the DDPM baseline, which again implies that the proposed CFRG noise schedule improves generation quality for low-frequency classes and confirms our findings.

\noindent{\bf Ablation on $\omega$ of Classifier Free Guidance.}
Classifier free guidance~\citep{ho2022classifier} is an important technique to trade-off between sample quality and diversity. We observe that other training techniques or experimental settings can influence the choice of the guidance scale $\omega$. Thus, for fair comparisons, the best guidance scale $\omega$ is individually selected for all the comparison models. We ablate the effects of $\omega$ in the following settings on ImageNet-LT with the image size of $32 \times 32$:
\begin{itemize}[leftmargin=1.0cm]
    \item Comparison between the DDPM models and models trained with our CFRG noise schedule.
    \item Comparison between CBDM models and models trained with our CFRG noise schedule.
\end{itemize}

Our experimental results are drawn on Figure~\ref{fig:guidance_scale}. An interesting phenomenon is that CBDM models require a larger guidance scale $\omega$. The CBDM method randomly and uniformly samples labels for current batch data. Then the ground truth labels and the sampled labels are both used in the training optimization with a weighted sum manner. As a result, there can be a negative effect on the alignment between image features and class embeddings. Then a larger $\omega$ can be a compensation for generation quality. Moreover, we also conclude that the larger image size demands a larger guidance scale $\omega$. with $64 \times 64$ image size, the best $\omega$ for our models and DDPM models is around 0.8 while around 0.3 for an image size of $32 \times 32$. Nevertheless, with the proper $\omega$ values for all of the models, we achieve much better results with the proposed CFRG noise schedule. 

\begin{table*}[t]
    \centering
    \caption{
        \textbf{Evaluation of image generation task on CIFAR-100-LT and ImageNet-LT.}
    }
    \label{tab:sota_cifar100}
    \resizebox{1.0\linewidth}{!}
    {
        \begin{tabular}{cclcccc}
            \toprule
            Dataset  & IF & Method & FID($\downarrow$) & Recall($\uparrow$) & $F_{s}(\uparrow)$ & $F_{1/s}(\uparrow)$ \\
            \midrule
            \multirow{8}{*}{CIFAR-100-LT} & \multirow{8}{*}{100}    &DDPM &7.38 &0.52 &0.85 &0.88         \\
                      & &+RS       &10.50  &0.49 &0.65 &0.83 \\ 
                      & &+SQRT-RS  &9.72   &0.47 &0.66 &0.83 \\
                      & &+DiffAug  &9.19   &0.47 &0.88 &0.86 \\  
                      & &+ADA      &6.16   &0.57 &0.91 &0.90 \\
                      & &+CBDM\&ADA &5.81   &0.57 & 0.91 & 0.90 \\
                      \cmidrule{3-7}
                      & &\textbf{CFRG(Ours)} &6.62(\textbf{-0.76}) &0.58(\textbf{+0.06}) &0.88(\textbf{+0.03}) &0.91(\textbf{+0.03}) \\
                      & &+CBDM\&ADA    &\textbf{5.14(-2.24)} &\textbf{0.60(+0.08)} &\textbf{0.91(+0.06)} &\textbf{0.93(+0.05)} \\
            \midrule
            \multirow{4}{*}{\shortstack{ImageNet-LT \\$32 \times 32$}}& \multirow{4}{*}{256} & DDPM     &3.09 &0.65 &0.76 &0.80  \\
                                    & &+CBDM &2.90 &0.65 &0.76 &0.79  \\
                                    \cmidrule{3-7}
                                    & &\textbf{CFRG(Ours)} &2.62(\textbf{-0.47}) &0.65(+0.0) &0.77(\textbf{+0.01}) &0.80(+0.0) \\
                                    & &+CBDM   &\textbf{2.33}(\textbf{-0.76}) &0.66(\textbf{+0.01}) &0.77(\textbf{+0.01}) &0.80(+0.0) \\
            \midrule
            \multirow{2}{*}{\shortstack{ImageNet-LT \\$64 \times 64$}}& \multirow{2}{*}{256} & DDPM     &6.02 &0.56 &0.71 &0.78  \\
                                    & &\textbf{CFRG(Ours)} &\textbf{5.50}(\textbf{-0.52}) &\textbf{0.57}(\textbf{+0.01}) &0.71(+0.0) &\textbf{0.79(+0.01)} \\
            \bottomrule
        \end{tabular}
    }
    \label{tab:sota_cifar100lt_imagenetlt}
    \vspace{-0.1in}
\end{table*}

\begin{table*}[tb!]
    \centering
        \caption{\textbf{Our CFRG models benefit image classification on imbalanced data, \textit{i.e.}, CIFAR-100-LT.}}
        \setlength{\tabcolsep}{0.23in}
         {
        \begin{tabular}{ccccc}
        \toprule
        Gen. Models &\multicolumn{2}{c}{IF=100} &\multicolumn{2}{c}{IF=200} \\ 
        \midrule
         &FID($\downarrow$) &Cls. Accuracy($\uparrow$) &FID($\downarrow$) &Cls. Accuracy($\uparrow$) \\
        \midrule
        - (Baseline) &-    &41.32  &- &37.09\\
        \midrule
        DDPM           &7.38 &46.58 &8.25 &42.79\\
        \midrule
        \textbf{CFRG(Ours)}         &6.62 &48.28  &7.46 &44.87\\
        +CBDM \& ADA &5.14 &\textbf{50.54(+9.22)} &5.89 &\textbf{46.26(+9.17)}\\
        \bottomrule
       \end{tabular}
       }
       \label{tab:cfg_cls}
       \vspace{-0.1in}
\end{table*}

\subsection{Image Generation}
\label{sec:generation}
To validate the effectiveness of the proposed class-frequency guided (CFRG) noise schedule on image generation task, we build several baselines based on DDPM~\citep{ho2020denoising} including re-sampling~\citep{mahajan2018exploring}, SQRT-resampling~\citep{mahajan2018exploring}, augmentation-based methods~\citep{karras2020training, zhao2020differentiable}, and class-balanced diffusion model~\citep{qin2023class}. We apply the CFRG noise schedule to DDPM and conduct experiments on the popular benchmarks suffering from the data imbalance issue, \textit{i.e.}, CIFAR-100-LT, and ImageNet-LT.

\noindent{\bf Image Generation Comparisons on CIFAR-100-LT.}
The experimental results on CIFAR-100-LT are listed in Table~\ref{tab:sota_cifar100lt_imagenetlt}.
We observe that the most classical method to deal with data imbalance, \textit{i.e.}, data resampling, is not effective for the image generation task with diffusion models. The class-balance resampling and SQRT resampling achieve 10.50 FID and 9.72 FID separately, which are even inferior to the DDPM baseline (7.38 FID). The ADA method, based on data augmentation, can significantly improve image generation performance, decreasing 7.38 to 6.16 in terms of FID. Currently, CBDM combined with the ADA method achieves the best performance on both generation quality and diversity. 
To show the effectiveness of our CFRG noise schedule, we apply our method to the DDPM baseline without extra techniques, achieving 6.62 FID and outperforming the baseline by 0.76 FID.
Plugging our CFRG noise schedule to CBDM and ADA methods, we boost the generation performance, largely surpassing the DDPM baseline model by 2.24, 0.08, 0.06, 0.05 in terms of FID, Recall, $F_{s}$, and $F_{1/s}$ evaluation metrics. The experimental results manifest the effectiveness and flexibility of our proposed CFRG noise schedule for diffusion models.

\noindent{\bf Image Generation Comparisons on ImageNet-LT.}
To demonstrate the generality of our CFRG noise schedule, we evaluate our models on the more challenging ImageNet-LT dataset. We conduct experiments with both $32 \times 32$ and $64 \times 64$ image sizes.
As shown in Table~\ref{tab:sota_cifar100lt_imagenetlt}, we achieve 2.62 FID and outperform the DDPM baseline by 0.47. Built on the CBDM method, our model obtains 2.33 FID, surpassing the DDPM baseline by 0.76. With an image size of $64 \times 64$, the proposed CFRG noise schedule consistently achieves much better performance than baselines.

\subsection{Image Classification with Generated Images}
\label{sec:image_classification}
To illustrate that the generative models can benefit downstream tasks, we evaluate the generated images by DDPM and our CFRG models on long-tailed recognition with the CIFAR-100-LT dataset.

\noindent{\bf Implementation Details.}
The experimental settings from previous work~\citep{cao2019learning, cui2022reslt} are deployed.
We randomly crop a $32 \times 32$ patch from the original image or its horizontal flip with 4 pixels padded on each side and normalize the pixel values into [0,1].
The ResNet-32 is used as the backbone network for all experiments. SGD optimizer with momentum 0.9 is adopted.
We train all models for 200 epochs with the cross-entropy loss. The initial learning rate is set to 0.1 and the first five epochs
are trained with the linear warm-up. The learning rate decays at the 160 and 180 epochs by 0.1. The
batch size is 128 and the weight-decay is 5e-4.

\noindent{\bf Experimental Results Comparisons.}
The experimental results are summarized in Table~\ref{tab:cfg_cls}.
We observe that generated data by diffusion models can significantly improve the classification performance on imbalanced data. With cross-entropy loss, the ResNet-32 model achieves 41.32\% and 37.09\% top-1 accuracy on CIFAR-100-LT with imbalance factor (IF) 100 and 200 respectively. Applying the same training settings, the model trained with generated data by DDPM achieves 46.58\% and 42.79\% top-1 accuracy, outperforming the baseline by 5.26\% and 5.7\%. Combining with ADA and CBDM methods, our CFRG model generates high-quality images. With these images, the ResNet-32 model boosts the classification accuracy to 50.54\% and 46.26\%, largely surpassing the baseline by \textbf{9.22\%} and \textbf{9.17\%} individually.

\begin{table}[t]
\caption{\textbf{The potential of CFRG noise schedule on the Stable Diffusion model.}}
\setlength{\tabcolsep}{0.3in}
\begin{tabular}{cc}
        \toprule
        Method  &FID($\downarrow$) \\
        \midrule
        DDPM      &22.84  \\
        w/ \textbf{CFRG(Ours)}    &\textbf{20.55(-2.29)}  \\
        \bottomrule
\end{tabular}
 \label{tab:kmeans}
 \vspace{-0.1in}
\end{table}

\begin{table}[t]
\setlength{\tabcolsep}{0.3in}
\caption{\textbf{Extension of CFRG to flow matching.}}
\begin{tabular}{cc}
     \toprule
     Method& FID  \\
     \midrule
     Flow Matching &11.05 \\
     w/ \textbf{CFRG}          &\textbf{10.60} \\
     \bottomrule
\end{tabular}
\label{tab:cfrg_flowmatching}
\vspace{-0.1in}
\end{table}

\subsection{Text-to-Image Generation}
\label{sec:text_to_image}
The Stable Diffusion (SD) model~\citep{rombach2022high} applies the mechanism of diffusion process to the latent feature space, largely reducing the computational cost.  It is trained on large-scale image-text pairs data, enabling text-to-image generation. Besides, the Stable Diffusion model can also secretly be an image classifier~\citep{li2023your}. As shown in previous work~\citep{Cui_2024_CVPR}, its zero-shot accuracy on CIFAR-100 is extremely imbalanced, which implies the large-scale training data also potentially suffers from the data imbalance issue.

Despite the reduced computational cost with latent features, training the Stable Diffusion model on large-scale vision-language data still requires large amounts of GPUs. To demonstrate the potential of our CFRG method on the Stable Diffusion model~\citep{rombach2022high}, we conduct simulated experiments on text-to-image generation with CIFAR-100-LT. The image encoder and text encoder are well-aligned in CLIP models~\citep{radford2021learning}. We thus use the features from the image encoder to represent their text embeddings. Then the text embeddings are adopted as the textual conditional inputs for diffusion models. At inference, we apply the template ``A photo of \{class name\}'' as text descriptions to generate 50,000 images. 
The experimental results are summarized in Table~\ref{tab:kmeans}.
The FIDs for both DDPM and our CFRG models are much higher than models trained with accurate labels, which implies the importance of high-quality data annotations. Nevertheless, under this text-to-image setting, our CFRG model still achieves better generation quality, confirming its great potential for the Stable Diffusion model. 

\subsection{Extension of CFRG to Flow Matching}
The objective of flow matching~\citep{lipman2022flow} still suffer from insufficient training in low-density regions ($q(x_{1} \approx 0)$):
\begin{eqnarray}
    v_{t}^{'} &=&  x_{1} - (1-\sigma_{min}) x_{0} \\
    \mathcal{L}_{CFM} &=& \mathbb{E}_{t, q(x_{1}), p(x_{0})} \left| v_{t}(\phi_{t}(x_{0})) \!-\! v_{t}^{'} \right|^{2}, 
\end{eqnarray}
where  $\phi_{t}(x_{0}) = (1-(1-\sigma_{min})t) x_{0} + tx_{1}$.

To mitigate this issue, we adapt the CFRG to flow matching and apply a class-wise $\sigma_{min}$: classes with higher frequency apply a smaller $\sigma_{min}$. 
We follow the open-sourced code from Facebook Research to train models for 3000
epochs and only replace the dataset as CIFAR-100-LT with an imbalance ratio of 100. Hyparameters are all set as default. 
The experimental results are summarized in Table~\ref{tab:cfrg_flowmatching}. After applying the CFRG strategy, the model performance is enhanced, demonstrating that our CFRG can also generalize to flow matching methods.

\section{Conclusion}
In this paper, we investigate the relationships between class frequency and the multi-scale noise schedule within diffusion models. Based on the observation that the low-frequency classes still suffer from large low-density regions and the high-frequency classes often dominate the estimated score space, a class-frequency guided (CFRG) noise schedule is introduced, which constrains that the noise scales are inversely related to class frequency. The effectiveness of our approach is confirmed via image generation, image classification, and text-to-image tasks.

\backmatter

\section*{Declarations}
\begin{itemize}
\item \textbf{Funding.} This research was funded by the National Natural Science Foundation of China (NSFC).
\item \textbf{Conflict of interest.} The authors declare that they have no conflict of interest.
\item \textbf{Data availability.} The datasets used in this work are all publicly available.
\end{itemize}

\begin{appendices}
\section{Proofs}
\subsection{The Relationship Between Score Function and Predicted Noise in DDPM}
\label{sec:score_function_ddpm}

\noindent{\bf Step-1}. DDPM~\citep{ho2020denoising} models the forward diffusion process as a Markov chain:
\begin{equation}
    q(x_{t}|x_{0}) = \mathcal{N}(x_{t}; \sqrt{\bar \alpha_{t}} x_{0}, (1-\bar \alpha_{t}) \mathbf{I}),
    \label{eq:markov_forward}
\end{equation}
where $x_{0} \sim q(x_{0})$ is the data distribution. $x_{t}$ represents noisy samples at time-step $t$, $\bar \alpha_{t}=\prod_{i=1}^{t} \alpha_{i}$, $\alpha_{i}=1-\sigma_{i}$, $\sigma_{1}, \sigma_{2},..., \sigma_{T}$ is the mutli-scale noise schedule~\eqref{eq:ddpm_noisy_schedule}.

\noindent{\bf Step-2}. The reverse diffusion process can also be considered as a Markov chain. With the Bayesian rule, forward diffusion process posteriors could be tractable:
\begin{eqnarray}
    q(x_{t-1}|x_{t},x_{0}) &=& \mathcal{N}(x_{t-1}; \tilde{\mu}_{t}(x_{t}, x_{0}), \tilde{\beta}_{t} \mathbf{I}), \\
    \tilde{\mu}(x_{t}, x_{0})&=&\frac{\sqrt{\bar \alpha_{t-1}} \sigma_{t}}{1-\bar \alpha_{t}} x_{0} \!+\! \frac{\sqrt{\alpha_{t}}(1-\bar \alpha_{t-1})}{1-\bar \alpha_{t}} x_{t}, \nonumber\\
    \tilde{\beta}_{t}&=&\frac{1-\bar \alpha_{t-1}}{1-\bar \alpha_{t}} \sigma_{t}. \nonumber
\end{eqnarray}
with $x_{0}=\frac{1}{\sqrt{\bar \alpha_{t}}}(x_{t}-\sqrt{1-\bar \alpha_{t}} \epsilon_{t})$ by Eq.~\eqref{eq:markov_forward}, 
\begin{equation}
    q(x_{t-1}|x_{t},x_{0}) \!=\! \mathcal{N}(x_{t-1}; \frac{1}{\sqrt{\alpha_{t}}}(x_{t} \!-\! \frac{1-\alpha_{t}}{\sqrt{1-\bar \alpha_{t}}} \epsilon_{t}), \tilde{\beta}_{t}\mathbf{I}),
    \label{eq:markov_reverse}
\end{equation}
where $\epsilon_{t} \sim \mathcal{N}(\mathbf{0}, \mathbf{1})$.

\noindent{\bf Step-3}. Applying the Tweedie's formula to $x_{t}$ in Eq.~\eqref{eq:markov_forward},
\begin{equation}
    \sqrt{\bar \alpha_{t}} x_{0} = x_{t} + (1-\bar \alpha_{t}) \nabla_{x_{t}} p(x_{t}), 
    \label{eq:tweedie}
\end{equation}
Combining Eqs.~\eqref{eq:markov_forward} and~\eqref{eq:tweedie}, we derive the following equation,
\begin{eqnarray}
    \sqrt{\bar \alpha_{t}} x_{0} + \sqrt{1-\bar \alpha_{t}} \epsilon_{t} &=& \sqrt{\bar \alpha_{t}} x_{0} - (1-\bar \alpha_{t}) \nabla_{x_{t}} p(x_{t})  \nonumber\\
    \Rightarrow \quad \nabla_{x_{t}} p(x_{t}) &=& \frac{-\epsilon_{t}}{\sqrt{1-\bar \alpha_{t}}},
    \label{eq:data_score_ddpm}
\end{eqnarray}
where $\epsilon_{t} \sim \mathcal{N}(\mathbf{0}, I)$.

Combining Eq.~\eqref{eq:data_score_ddpm} and Eq.~\eqref{eq:markov_reverse}, Eq.~\eqref{eq:markov} is established for DDPM, implying the importance of accurate score estimation for sampling at inference.

\begin{table*}[t]
    \centering
    \caption{
        \textbf{Experimental Configurations.}
    }
    \label{tab:exp_config}
    \resizebox{1.0\linewidth}{!}
    {
        \begin{tabular}{cccccccccc}
            \toprule
            Dataset &Method  &$\sigma_{T}^{min}$ &$\sigma_{T}^{max}$ &$\sigma_{1}$  &$\gamma$ &Lr &Batch size &Num GPUs & Train iters \\
            \midrule
            \multirow{5}{*}{CIFAR-100-LT} &DDPM                &0.02 &0.02 &1e-4 &-     &2e-4 &64  &4 &30w \\
                                        &+CBDM               &0.02 &0.02 &1e-4   &-     &2e-4 &128 &4 &30w \\
                                        &+CBDM \& ADA           &0.02 &0.02 &1e-4   &-     &2e-4 &128 &4 &50w \\ 
                                        \cmidrule{2-10}
                                        &CFG(Ours)          &0.02 &0.04 &1e-4   &0.999 &2e-4 &64  &4 &30w \\
                                        &+CBDM \& ADA &0.02 &0.04 &1e-4   &0.999 &2e-4 &128 &4 &50w \\
            \midrule
            \multirow{5}{*}{\shortstack{ImageNet-LT \\$32 \times 32$}} &DDPM               &0.01 &0.01 &1e-4   &-     &2e-4 &512   &4 &50w \\
                                        &+CBDM               &0.01 &0.01 &1e-4   &-     &2e-4 &1024  &4 &50w  \\
                                        \cmidrule{2-10}
                                        &CFG(Ours)          &0.01 &0.02 &1e-4   &0.995 &2e-4 &512   &4 &50w  \\
                                        &+CBDM     &0.01 &0.02 &1e-4   &0.995 &2e-4 &1024  &4 &50w \\ 
            \midrule
            \multirow{2}{*}{\shortstack{ImageNet-LT \\$64 \times 64$}} &DDPM               &0.01 &0.01 &1e-4   &-     &2e-4 &512   &8 &50w \\
                                        &CFG(Ours)          &0.01 &0.02 &1e-4   &0.995 &2e-4 &512   &8 &50w  \\
            \bottomrule
        \end{tabular}
    }
\end{table*}
\begin{table*}[t]
    \centering
    \caption{
        \textbf{More results on image generation task with CIFAR-100-LT.}
    }
    \label{tab:sota_cifar100_if200}
    \resizebox{1.0\linewidth}{!}
    {
        \begin{tabular}{cclcccc}
            \toprule
            Dataset  & IF & Method & FID($\downarrow$) & Recall($\uparrow$) & $F_{s}(\uparrow)$ & $F_{1/s}(\uparrow)$ \\
            \midrule
            \multirow{3}{*}{CIFAR-100}& \multirow{3}{*}{200} & DDPM       &8.25 &0.50 &0.86 &0.82  \\
            \cmidrule{3-7}
                                      & &\textbf{CFG(Ours)} &7.46(\textbf{-0.79}) &0.56(\textbf{+0.06}) &0.86(+0.00) &0.89(\textbf{+0.07}) \\
                                      & &+CBDM\&ADA    &\textbf{5.89(-2.36)} &\textbf{0.58(+0.08)} &\textbf{0.88(+0.02)} &\textbf{0.90(+0.08)} \\
            \bottomrule
        \end{tabular}
    }
\end{table*}

\subsection{Low-density Regions Lead to Inaccurate Score Estimation for DDPM.}
\label{sec:inaccurate_score_ddpm}

The optimization of DDPM~\citep{ho2020denoising} is driven by a variational bound on the negative log-likelihood. Specifically, the objective is written as follows:
\begin{eqnarray}
   \mathcal{L}_{T} &=& D_{KL}(q(x_{T}|x_{0})\| p(x_{T})) \nonumber\\
   \mathcal{L}_{t-1}&=&\sum_{t>1}D_{KL}(q(x_{t-1}|x_{t},x_{0}) \| p_{\theta}(x_{t-1}|x_{t})) \nonumber\\
   \mathcal{L}_{0} &=& \log p_{\theta}(x_{0}|x_{1}) \nonumber\\
   \mathcal{L}_{all} &=& \mathbb{E}_{q} \left[\mathcal{L}_{T}  + \mathcal{L}_{t-1} + \mathcal{L}_{0} \right],
   \label{eq:ddpm_loss}
\end{eqnarray}
where $p_{\theta}(x_{0:T})$ represents the reverse diffusion process, $p(x_{T}) \sim \mathcal{N}(\mathbf{0}, \mathbf{I})$.

\noindent{\bf For $t>1$}, 
\begin{eqnarray}
    \mathcal{L}_{t-1}&=& D_{KL}(q(x_{t-1} | x_{t}, x_{0}) \| p_{\theta}(x_{t-1}|x_{t})) \\
                     &=& \frac{1}{2 \tilde{\beta}_{t}^{2}} \|\mu_{p_{\theta}} -\mu_{q} \|_{2}^{2} \nonumber\\
                     &=&  \frac{1}{2 \tilde{\beta}_{t}^{2}} \| \frac{1}{\sqrt{\alpha_{t}}}(x_{t} - \frac{1-\alpha_{t}}{\sqrt{1-\bar \alpha_{t}}} \epsilon_{\theta(x_{t}, t)}) \nonumber\\
                     &-&\frac{1}{\sqrt{\alpha_{t}}}(x_{t} - \frac{1-\alpha_{t}}{\sqrt{1-\bar \alpha_{t}}} \epsilon_{t})  \|_{2}^{2} \nonumber\\
                     &=& \frac{(1-\alpha_{t})^{2}}{2 \tilde{\beta}_{t}^{2} \alpha_{t}} \|\frac{-\epsilon_{\theta}(x_{t},t)}{\sqrt{1-\bar \alpha_{t}}} \!-\! \nabla_{x_{t}} p(x_{t}) \|_{2}^{2}, \nonumber
    \label{eq:ddpm_kl}
\end{eqnarray}
where the learned score function $s_{\theta}=\frac{-\epsilon_{\theta}(x_{t},t)}{\sqrt{1-\bar \alpha_{t}}}$.

With Eq.~\eqref{eq:ddpm_kl} and Eq.~\eqref{eq:ddpm_loss}, we conclude that the DDPM model suffers from insufficient training for low-density regions and thus leads to more inaccurate score estimation for low-frequency classes.

\begin{algorithm*}[h]
    \caption{Quantity of low-density regions.}
    \label{algorithm:density_analysis}
    \begin{algorithmic}
    \State Step 1: we assume pixel $p \sim \mathcal{N}(\mu_{c}^{p}, \sigma_{c}^{p})$ for each sample in class c.
    \State Step 2: generate 500 samples for each class, denoted by $\mathcal{D}_{bal}$
    \State Step 3: kde = KernelDensity(kernel='Gaussian', banwidth=0.5).fix(CLIP($\mathcal{D}_{bal}$))
    \State Step 4: derive $\mathcal{D}_{cfrg}$ and $\mathcal{D}_{ddpm}$ with CFRG and DDPM models taking $\mathcal{D}_{bal}$ as inputs.
    \State Step 4: log\_p\_cfrg = kde.score\_samples(CLIP($\mathcal{D}_{cfrg}$))
    \State Step 5: log\_p\_ddpm = kde.score\_samples(CLIP($\mathcal{D}_{ddpm}$))
    \State Step 6: low\_density\_cfrg = np.mean(log\_p\_cfrg $\leq \delta$ )
    \State Step 7: low\_density\_ddpm = np.mean(log\_p\_ddpm $\leq \delta$)
    \end{algorithmic}
\end{algorithm*}

\section{Settings}
\label{sec:exp_config}
\subsection{Experimental Settings}
\noindent{\bf Datasets.} 
CIFAR-100 has 60,000 images –-- 50,000 for training and 10,000 for validation with 100 categories. To illustrate the importance of class frequency for noise schedule designing in diffusion models, we use the long-tailed version of CIFAR datasets with the same setting as those used in \citep{cao2019learning, cui2019class}. The degree of data imbalance is measured by an imbalanced factor $\frac{N_max}{N_min}$, where $N_{max}$ and $N_{min}$ are the most and the least class frequency in the dataset. In addition, we also conduct experiments on the more complex data distribution --- ImageNet-LT~\citep{liu2019large}. ImageNet-LT is a long-tailed version of the ImageNet dataset~\citep{russakovsky2015imagenet} by sampling a subset following the Pareto distribution with a power value of 6. It contains 115.8K images from 1,000 categories, with class cardinality ranging from 5 to 1,280.

\noindent{\bf Implementation Details for the Diffusion Model.}
Observing that the default hyper-parameters in DDPM also achieve best results on CIFAR-100-LT,  we adopt the same training configurations as the baseline model, \textit{i.e.}, DDPM~\citep{ho2020denoising}. $\sigma_{1}=1e-4$ and $\sigma_{T}=0.02$ with $T$ = 1, 000 are set for the noise schedule. We optimize the network with an Adam optimizer whose learning rate is 0.0002 after 5,000 iterations of warmup. On ImageNet-LT, we select the best hyper-parameters for the baseline model. $\sigma_{T}=0.01$ is used. Image sizes of $32 \times 32$ and $64 \times 64$ are applied to training optimization and evaluation. Considering that the size and semantic complexity of the datasets vary greatly, we choose appropriate epochs for each dataset. 4 Nvidia GeForce 3090 GPUs are used for training. 

\noindent{\bf Evaluation Metrics.}
Models trained with our CFG noise schedule and the corresponding baseline models are compared with respect to both the generation diversity and fidelity via Frechet Inception Distance (FID)~\citep{heusel2017gans}, Recall~\citep{kynkaanniemi2019improved} and $F_{\beta}$~\citep{sajjadi2018assessing}. The Recall and $F_{\beta}$ are measured using features from Inception-v3 that are pre-trained on ImageNet. Following the practice of CBDM, we take $K=5$ for the Recall, $1/8$ and $8$ for the threshold in $F_{\beta}$, and $20$ times of class number as the clustering number of $F_{\beta}$ to capture the inner class variance. In the evaluation, we use all the real images from the training set of CIFAR-100 and ImageNet rather than the imbalanced training data during training optimization. Classifier-free guidance~\citep{ho2022classifier} is applied to both our models and baseline models. For fair comparisons, the guidance strength $\omega$ is carefully tuned for all of these models.

\noindent{\bf Experimental Settings.}
Our experimental settings are listed in Table~\ref{tab:exp_config}.
Except on ImageNet-LT with the image size of $64 \times 64$, we conduct experiments on 4 Nvidia GeForce 3090 GPUs.
Our method is more efficient than CBDM. CBDM requires additional regularization batch data for each training iteration. However, there is no extra computational cost for our approach compared to the DDPM baseline. On CIFAR-100-LT, it takes around 1 day for 300,000 training iterations of our method. Equipped with CBDM and ADA, around 3 days are required to complete the 500,000 training iterations.
On ImageNet-LT with the image size of $32 \times 32$, it takes around 3 days for 500,000 training iterations of our method.  With the image size of $64 \times 64$, 5 days are needed to finish 500,000 training iterations using 8 Nvidia GeForce 3090 GPUs. Under inference, it takes 1 day to generate 50,000 $32 \times 32$ images while 4 days to generate 50,000 $64 \times 64$ images with 1 Nvidia GeForce 3090 GPU.

\noindent{\bf More Details on the Case Study.}
For CIFAR-100-LT, we divided the classes into ``Many''($\ge 100$ images), ``Medium'' ($\ge 20$ images), and ``Few'' ($<20$ images) three groups according to their class frequency.  On ImageNet-LT, the classes with ``$\ge 500$ images'', ``$\ge 50$ images'' and ``$<50$ images'' are categorized into ``Many'', ``Medium'', and ``Few'' groups respectively.

\subsection{More Experimental Results}
\label{sec:more_results}
We conduct experiments on CIFAR-100-LT with an imbalance factor of 200. The results are listed in Table~\ref{tab:sota_cifar100_if200}. With the proposed class-frequency guided (CFG) noise schedule, we achieve 7.46 FID, surpassing the DDPM by 0.79. Moreover, the CFG noise schedule is orthogonal to previous methods, like ADA and CBDM. Equipped with CBDM and ADA methods, we achieve 5.89 FID, significantly outperforming the DDPM by 2.36.

\subsection{Density Analysis Details}
The quantity of low-density regions in Table~\ref{tab:density_error} is measured by algorithm~\ref{algorithm:density_analysis}.

\subsection{Code}
\label{sec:code}
Our code is provided at \url{https://drive.google.com/file/d/1kNb-DSOQBlpp8330_PKgRB_FAGOe79i_/view?usp=sharing}.




\end{appendices}


\bibliography{sn-bibliography}

\end{document}